# Poetry "resurrection" from LLMs

**Vincent Cellucci, Peter van der Putten**
LIACS, Leiden University
Leiden, the Netherlands
{v.a.cellucci, p.w.h.van.der.putten}@liacs.leidenuniv.nl

### Abstract

In an era where linguistic creativity is challenged by algorithmic authorship, we pose a provocative question: can artificial intelligence (AI) write an infinite poem? We set up two experiments to investigate the quantitative and qualitative output produced by AI utilizing couplets that evidence a poetic relationship between end words, the only variation in a strict anaphoric, litany structure. Our results show that while AI can generate interesting pairings, it still falls short of human-level creativity and understanding of poetic relationships. We offer commentary and a brief survey of the field of computational poetry experiments, tracing back authorial intention and distinctions of the "natural" and "artificial" poetry, as well as justifying why couplet form is suited to testing AI's poetic output. Experiment 1 relies on manual prompting and co-creation to extend the original poem, which is developed into an animated webpage installation and interactive artwork. Experiment 2 reliably achieves automated output reaching the 1000 benchmark via an alternative method of using an LLM in combination with semantic similarity, of which Method #2 demonstrates a higher capacity for semantic and quantitative poetic output. These experiments highlight the limitations of current LLMs' generation capacity at the word level but also suggest that poetry will also respond and revive by incorporating AI workflows.

## Introduction

Large language models (LLMs) as a method of generative artificial intelligence (AI) promise an almost infinite capacity for generative output—a crisis in its own right for many authors and creatives. As the lead author is a poet focusing on computational creativity enhancing traditional poetic expression techniques, our research team decided to test generative AI's ability to write an infinite poem.

This research uses an iterative design process based on two experiments and two iterations of the artwork *resurrection*. The two experimental artworks are offered to open a discussion of further methodologies and to improve automated poetic output as well as semantic relationships. We also provide reflection on the crisis of generative AI's threat to creative output. This crisis harkens back to what literature tries to overcome, humanity's original crisis: mortality.

The advent of AI or LLMs, has democratized poetry generators, putting these capacities outside specialists in computer science and poetry. As frequent reviewers for design and media technology courses at Leiden and Delft University, we have certainly seen our fair share of AI-powered poetry student projects. While we are always game for fun interactive experiences with language, we must acknowledge that many projects hinge on poetry merely for a sense of profundity or future-telling, oracles resulting in predictions or fortune-cookie-like texts printed on slips of paper. Others use AI poetry generation as an atmosphere for sensory-based experiences either visual projections or aural playback from recordings.

We consider all these experiences primary research to inform our empirical understandings of how AI has changed the way we experience and create poetry. They inspire us to reconsider our book-centric notions of experiencing poetry and the usage (or refusal) of technology in the writing process, leading us to wonder what could be discovered when we try to apply a more technical understanding of poetry to make an AI-generated poetry experience that incorporates interactivity and tests the limits and extent of poetic output.

We approach this question by designing and exploring a form of infinite poem: an open-ended generative poem. In this paper, "infinite" is used conceptually to describe a system capable of unbounded extension rather than literal infinity. Our specific angle is to study lexical-level creativity and poetics under extreme formal constraint. More concretely, this paper presents the following contributions:

- A constrained-form experimental framework and testbed for testing lexical-level poetic generation based on anaphoric couplets
- A comparative pipeline combining vector similarity and LLM pairing for constrained poetry
- A small-scale quantitative evaluation of, and an artist-led, subjective reflection on the poetic pairings
- Two artistic-research installations demonstrating long-form generative poetic accumulation

## Related Work

First, let us give a brief overview of the literature pertaining specifically to computational poetry experiments. In 1996, Charles Hartman published *Virtual Muse: experiments in computer poetry*, which has become a milestone text in the field as it briefly outlines the history of computational poetry experimentation that traces to a similar origin publication for a solely-computer written book, *The Policeman's Beard is Half-Constructed* (1984) by a program called Racter coded by William Chamberlain and Thomas Etter capable of producing an array of output, even conversation. [1] More recently, poets and programmers Lillian-Yvonne Bertram and Nick Montfort edited an anthology titled *Output*, which collects computer-generated texts of projects from 1953 to 2023. They have managed to identify and provide more information about earlier computational poetry projects. Such projects consist of programs to write: "love letters" (Christopher Strachey, 1953); screenplays for TV (Doug Ross and Harrison "Dit" Morse, 1960); stochastic texts (Theo Lutz, 1959); and poetry that imitates a literary style (R.M. Worthy, 1961). These examples detail the initial attempts at linguistically generating "profound serendipity." [2]

In 1962, Max Bense was likely the first scholar to differentiate between "natural" and "artificial" poetry. He supported his distinction in Hegelian terms of a "personal poetic consciousness" behind "natural" poetry, leading him to posit "artificial poetry can only have a material origin" where "all words are a priori equal" as opposed to "natural" poetry, which "privileges [*Vorzugsstellung*]" word classes for semantic reasons—ultimately facilitating the end goal of interpretation. [3] Interestingly enough, Bense does allow for humans to generate "artificial poetry" via "structural word programs" and "serial word sequences." [4] Bense cites concrete poetry and Stein as examples. He also brings up the term "cybernetic" as a branch of artificial poetry for when "the selection (i.e., the writing) is carried out by the machine"—leading one to believe this is how he would characterize our current use of LLMs. Bense furthers his distinction by advising usage of the term "texts" to indicate artificial language output; with this assertion, as German scholar, Hannes Bajohr points out as context to his translation, Bense introduces the problem of intention, the same problem society is confronted with today given the rampant use of AI/LLMs disrupting our fixed notions of authorship, aesthetics, interpretation, and meaning. [5]

The Introduction of *Output* contrasts the computational works presented in the anthology to the advent of LLMs and Gen AI, which is predominantly used for business or academic writing—used casually by anyone with access—while also noting "the production of computer-generated poetry and novels using LLMs has also swelled significantly." [6] Before this societal condition that has arisen relatively recently, this work was pursued by programmers who "directed their skills, in their free time, toward exploratory and artistic purposes." [7]

Poetry generation is a complex task, and there have been plenty of documented and undocumented poetry generators. One of the most cited interactive poetry generators in recent literature is *PoeTryMe*. [8] Written in JavaScript, *PoeTryMe* focused on generating poetry in various languages (e.g., English, Portuguese, or Spanish) from user-contributed seed words, although our assessment (which is limited by lack of access to the tool) is that the output is primarily related to structure, form and meter (lines, stanzas, and syllables), essentially making it look like a poem more than isolating to plumb single poetic techniques. [9] Notably, in a work from the same pre-LLM period using Markov chains, *Diamonds in Dystopia*, the authors refer to their original poem as the seed poem, so there is an intuitive, common vocabulary shared between people designing these computational poetry tools. [10]

Another poetry generator pertinent to our experimentation is the project *Nehovah*, which produces neologisms, or new words, from user-contributed source words, generating output by blending synonyms from Top Ten lists found online; interestingly enough, as a poet that prioritizes sound and image, the lead author disagrees with the assessment of the output, finding some of the example output classified as "worst," belong in "best" and vice versa. [11] This point is brought up mainly to reify the spectrum of associations and expectations related to poetry (with direct bearings on people's education, experience, and stylistic leanings—certainly natural biases for assessment of generated output).

Neither of these generators used LLMs, so we want to at least mention two admirable projects, *1 the Road* and *No solitude in the caves of spring*. [12,13], that rely upon recurrent neural networks, a precursor neural network method to modern-day transformer-based LLMs. These projects are forerunners for generative texts produced directly from neural networks into literary practices, but they were not intended as experiments targeting the science behind computational creativity used for poetry.

## Poetic Techniques: Couplet, Litany, Anaphora

To begin experimentation, we needed some strict poetic constraints to prevent open-ended exploration. Based on one of the lead author's experimental poems titled "resurrection," we decided to focus on one of the simplest forms: the couplet, whose closed form is "well-suited to express aphoristic wit." [14]

The use of couplets as units of composition in verse extends before 14C English, given Chaucer introduced the heroic couplet during this period with *The Canterbury Tales* (and the written form clearly derives from the oral tradition's employment of end-rhyme to aid memorization). Poet and critic Timothy Steele claims, "Poets in English have employed the heroic couplet...more frequently and variously than any other traditional rhymed form." [15] Presumably, this abundance would also be representative in the LLM's training data.

Sir Philip Sidney, in 16C, furthered our understanding of the couplet. In *The Countesse of Pembrokes Arcadia*, Sidney wrote of Pan and his companions "singing some short coplets, whereto the one halfe beginning, the other halfe should answere"—drawing out the relationship between the lines. [16] This relationship also contains an element of intensification or competition, "Having thus varied their songs and dances into diverse sorts of inventions; their last sport was one of them to provoke an other to a more large expressing of his passions." [17] Inheriting the competition of poetic expression intrinsic to this form, we naturally wanted to see if AI could do a better job at extending my poem (infinitely), which leads to the other operative poetic technique of litany.

| | |
|---|---|
| one more song<br>one more dance | one more storm<br>one more rebirth |
| one more drink<br>one more glance | one more boil<br>one more pour |
| one more scream<br>one more delight | one more dash<br>one more slather |
| one more float<br>one more night | one more rash<br>one more shakedown |
| one more stance<br>one more rant | one more martyr<br>one more father |

Figure 1. Sample of couplet form taken from the original, human-authored "resurrection."

The second poetic technique employed, litany, stems from "magical and liturgical origins," and can be found in abundance in the Bible, children's books, political speeches, and poems, leaving us "spell-bound" (similar to the current powers of AI). [18] It is a pattern we easily learn and adapt (one ostensibly simple enough for AI to recognize and learn as well), as its repetition combined with variation provides great rhetorical effect. Litany provides a poem structure highly suitable for quantitatively testing the generative output of AI.

To further limit the constraints of litany for this study, strict anaphora is used as a control because this technique consists of the exact repetition of the words at the beginning of the line, fixing the words that are not generated. Our study uses only two words, which repeat (endlessly) form the entire structure of the poem.

## The Poem: "resurrection" as a testbed for generative AI

The backdrop of this human-authored experimental poem relates to overcoming two crises (Hurricane Katrina and COVID); it is the origin of the experiments' simple structure formed from an anaphoric, litany-based on couplets with related end words. In layman's terms, it repeats the phrase "one more" followed by a new pair of words. Figure 1 demonstrates this constrained couplet form.

This form highlights the relationship between end words. This "relationship" is not as simple as it might appear. True, the poem initiates and builds its initial momentum and relationships with rhyme, but as later couplets evidence, the form does not require it nor depend on it being sustained with regularity. Also, mere antonyms are not enough to sustain the poem, even though they manifest readily as one of the most identifiable relationships for generating couplets. Other associations appear like images, connotations, or idiomatic ones. Note that the structure even has a second tier of opportunities to build relationships between couplet series. Even a small sample shows the endless capability of such wordplay from a human author. The poet could have kept going but instead stopped the original poem at 121 couplets, which was edited down from considerably more, proving this constrained form had high potential to help us research what creative relationships might exist.

The phrase "one more" is commonly shouted in jazz clubs of New Orleans as a plea for an encore. Written from the dramatic point of view of a dying person, this poem co-opts that language. It becomes a celebration of all of life, with no limits besides the formal constraint. Therefore, it becomes an even more fascinating linguistic experiment to give to a machine presumably surpassing the poet's capacity for counting, analysis, and recombination...while acknowledging this machine is without life experiences and incapable of the nostalgia ingrained in the tone of this poem.

### Experiment 1: couplet generation via manual prompting

Prioritizing couplets and litany led to a two-fold structure for experimentation which allowed for both the combined and separate examination of quality and quantity, a distinction particularly useful when working with generative AI. Our research questions could then be summarized as: 1) can LLMs produce large numbers of couplets without repetition (quantity) and 2) could the LLM produce couplets with a relationship (from the variety of possible relationships) that would satisfy a poet enough to incorporate them into the poem (quality).

Preliminary testing was done using a selection of the poem as training data on a variety of models to determine which one might give the best results. We ran some trials with open source LLMs (e.g., Llama 3.1). These were quite limited in their output from a creative perspective, evidencing more random words than identifiable relationships. Alternatively, we found DeepSeek-R1 to have a slightly higher edge in terms of creativity of the output, but it was not reliable in returning the quantity of results needed as the server would often be busy, allowing no results. (N.B DeepSeek-R1 only available for download and running locally in January 2025.) Therefore, we elected to use the GPT 4o model because initial tests demonstrated it was highly reliable and could return the most couplets and couplets with recognizable relationships between end words.

Once committed to GPT 4, we used the entire original poem as training data in my prompt, while asking it to generate batches of 100 couplets with no repetition "forming creative and surprising juxtapositions without repeating any end words." Surprisingly, no matter the number of requests and reprimands that emphasized "no repetition" of the end words, ChatGPT was incapable of following this mandate. In the worst case of returns, a subsample resulted in 21 out of 25 responses with exact repetition, in this case of the word "ember." Shocked by the redundancy of these prompting sessions, we started compiling a list of ChatGPT's favorite words: "ember, gossamer, mosaic, monsoon, nebula, reverie, ephemeral, and fjord" all had many occurrences, which we presume is a reflection of the training data (presumably 19th and 20th century novels in the public domain).

Because of the high quantity of repetition as well as the need for getting to a rudimentary benchmark, we limited exponential expectations for the combined written and generated couplets for the co-created poem, setting the goal of a total of 365 couplets total. Quantitatively, this figure is a bit arbitrary, except as a reflection of the state of the limited output possible for LLMs through prompting during this period, as we were hard-pressed using manual methods to achieve the additional 244 couplets.

Conceptually, 365 represents a year's worth of daily couplets to the poem, which relates to a potential outcome for an infinite AI-generated poem that can be updated daily online. This is a gesture inspired by Japanese artist On Kawara's "Today" series of simple black and white font date paintings, which profoundly reveal that the minimal job of the artist is to create something each day. If you extend this contemporary art definition of artmaking to poetry, then an AI capable of writing a new couplet each day is indeed a poet more than any qualitative analysis provides of the semantics behind the generated text. This mode of reasoning compounds post-modernism's upending questions about the roles of the artist with those that have been exaggerated by artificial intelligence's application in creative arts.

As a strategy to get richer diction and less repetitive output, we specified to mine rare or less frequently used words and specialized lexicons, with prompts related to science, space, health, medicine, law, etc. We even resorted to stripping away the structure of the poem and just asking for lists of "nouns that evoke vivid imagery, capturing the essence of the natural world, human experience, and abstract concepts", which were used to make a pleasing pairing relationship, the poet assuming creative control after attempting to cede it to the machine.

## Experiment 1: Results and Conclusions

Eventually, we were able to achieve 244 "generated" couplets from manual prompting sessions of ~1600 outputs (accepting 15% of entries). At this stage, the only way to reach the simplified benchmark goal (of 365 total) was by heavily relying on mediation and human–AI co-creation to combat severe issues of repetition and unidentifiable relationships between end words (randomness in pairings). Along with these two essential criteria, super cliché or too alike synonyms were additional semantic parameters that were reasons for excluding generated output. For example, there were many generated couplets like "one more atom / one more molecule" that, while appearing acceptable, have a more synonymous than poetic relationship, although one can detect occurrences of other poetic uses of language, like synecdoche. But too many formations that are similar are not acceptable, revealing the LLM is not capable of considering the whole dataset as intuitively as a human author. Poetry is about excluding and curating language as much as it is about simply generating it, so perhaps the evaluation and exclusion abilities of LLMs are something to investigate in the future.

Experiment 1 concluded with a usable artifact that pleased the poet, but the researcher inside him was disappointed, as the methodology was not repeatable nor achieving desired outcomes. The poet had backslid into curated, manual methods of creation, which allowed him a high degree of quality control and resulted in a traditional print publication for the poem. [19] Sure, he could claim that he added artificial intelligence to his process, a new tool for co-creation, but this usage certainly conforms to traditional notions of AI as a tool...[to] augment human creativity." [20] To grow as a researcher, he needed to increase his technological proficiency and methods and pursue less manual ones if he wanted to obtain goals in the form of increasingly higher benchmarks.

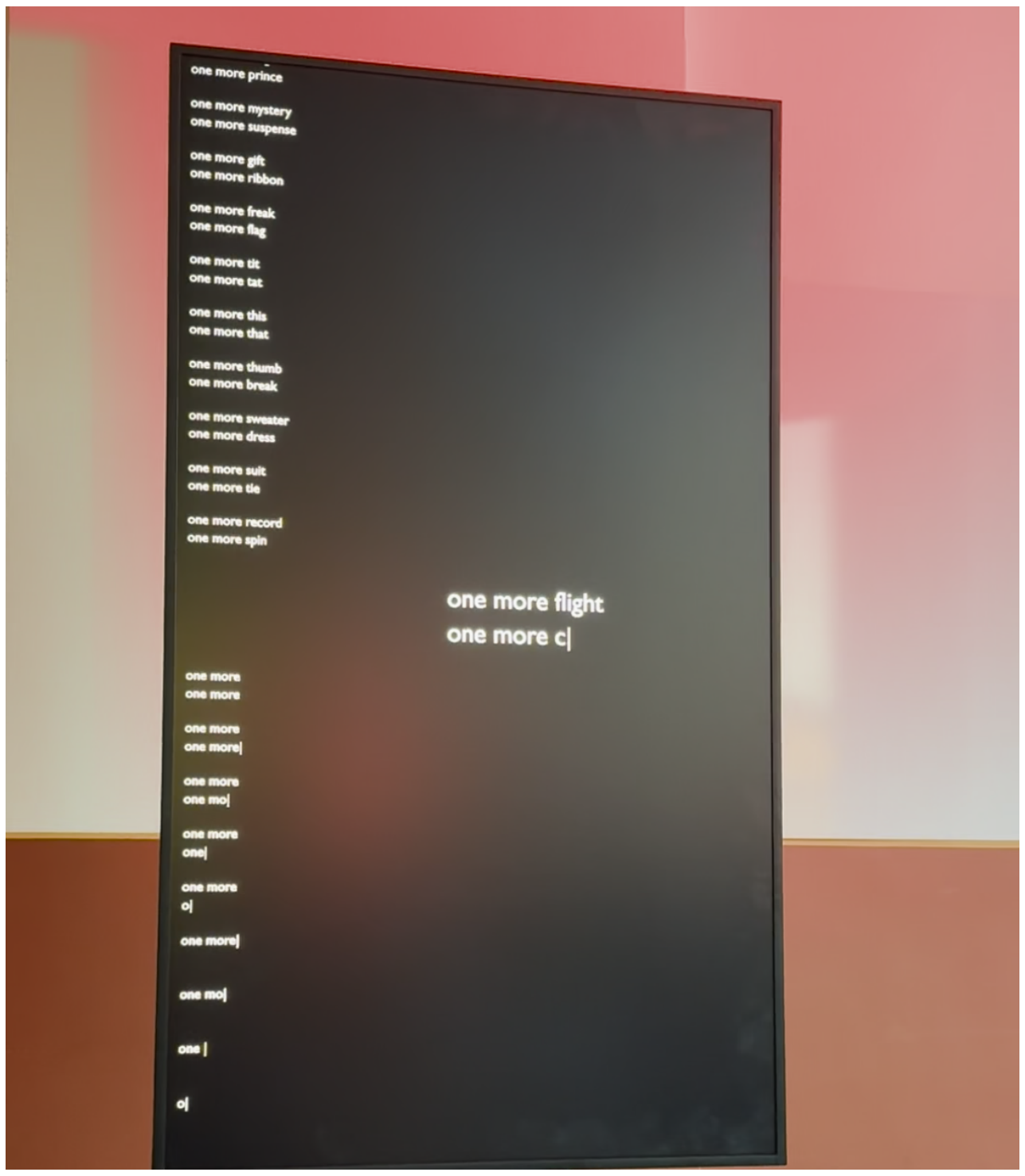

Figure 2. *resurrection*, animated web poem consisting of 365 human, AI-generated, and co-created couplets, Night of Discoveries, 2024.

## Artwork 1: *resurrection* animated web poem

For the September 2024, Night of Discoveries festival, the lead author exhibited a webpage of *resurrection* that animated the couplet generation (Figure 2). The artwork used portrait orientation to maximize the litany effect and a typewriter animation with a delay on the second line of the couplet so that the viewer perceives (a simulacra of) the machine thinking and composing a related end word. The animation was kept simple, co-opting On Kawara's minimal visual aesthetic with an added glitch art ending to provide some visual diversity and a payoff for anyone who lasts through the 20-minute cycle of the 365 couplets. The animation does not distinguish between human, AI-generated, or co-created couplets.

## Experiment 2: automated couplet generation

To continue this pursuit to reach some semblance of AI-assisted poetic infinity, we realized the need for more automated methods. We set a new modest, quantitative benchmark for surpassing 1,000 outputs.

The computational method that we deemed the most promising for exploration was combining LLMs and semantic similarity search. We used ChromaDB as a vector database, with sentence transformer embeddings (based on all-MiniLM-L6-v2) with Euclidean distance. [21] We loaded the database with a Scrabble wordlist[1] of 279,496 entries. In Python, we created a couplet generator that was capable of producing output from a two-step process that was interchangeable. We split the tasks of selecting a word and pairing the word. This way, we could test and compare the two methods. In Method #1: the first word is generated by AI (using OpenAI's API running GPT4 for model consistency with Experiment 1), and the paired second word is the $n$-th most similar word, with $n$ a random number between 50 and 150, so that a related word is selected without being too obvious. In Method #2, the first word is selected at random from Scrabble wordlist in ChromaDB, and the pairing (second word) is done using AI (GPT4).

We expected Method #2 to outperform on the qualitative task of creatively pairing the end words. But Method #1 stood a better chance of being much less repetitive and mining a richer vocabulary to draw from, therefore increasing the quantity.

## Experiment 2: Results and Conclusions

Initial testing and setting up the couplet generator proved competent in exceeding the 1000 output benchmark with regularity. A storage mechanism was added to the script to capture the couplet output in a comma-delimited .json (CSJ) to facilitate parsing and ultimately analysis.

Because of the automated prompting, repetition was less of a nuisance, but it was still an issue, especially in Method #1, which required us to employ an additional recursive script to filter out GPT's redundancies. In Method #2, this became less of an issue because GPT was pairing, yet there was still identifiable repetition of roughly 12.8% of the output's sample size. Nevertheless, both methods enabled automated generation of +1000 outputs.

Rather than keep pushing the quantitative limits of the methodology, we deemed it more valuable to return our focus to the creative (or at least related) language generation by exploring the data to qualitatively assess the output through the lens of a poet.

We analyzed a set of data manually for each method, developing a codified system to classify the most prevalent end word relationships generated by the computational pairings. These consisted of: word or word form similarity; sonic or visual rhyme, including words that both sounded or looked the same as well as alliterative pairings; synonyms; antonyms; cliché (not always a bad thing as this category was also used to include words commonly or culturally associated, including idiomatic or more verbal slang); synecdoche, a part to whole relationship; and unrelated. Concurrently, the poet indicated a "poetic" pairing—by design, a purely subjective decision—based on the binary criteria of whether it could be accepted, unedited, into his poem. Of course, there were multiple temptations for co-curation with minor tweaks, but he resisted, and these entries did not receive the "poetic" designation.

| Method | Not Poetic | Poetic |
| --- | --- | --- |
| # 1: GPT + CDB | 91% | 9% |
| #2: CDB + GPT | 47% | 53% |

Figure 3. Overall acceptance rate of output of 1000 generated couplets using Method #1 and a sample of 250 generated couplets from Method #2 (GPT4 + ChromaDB).

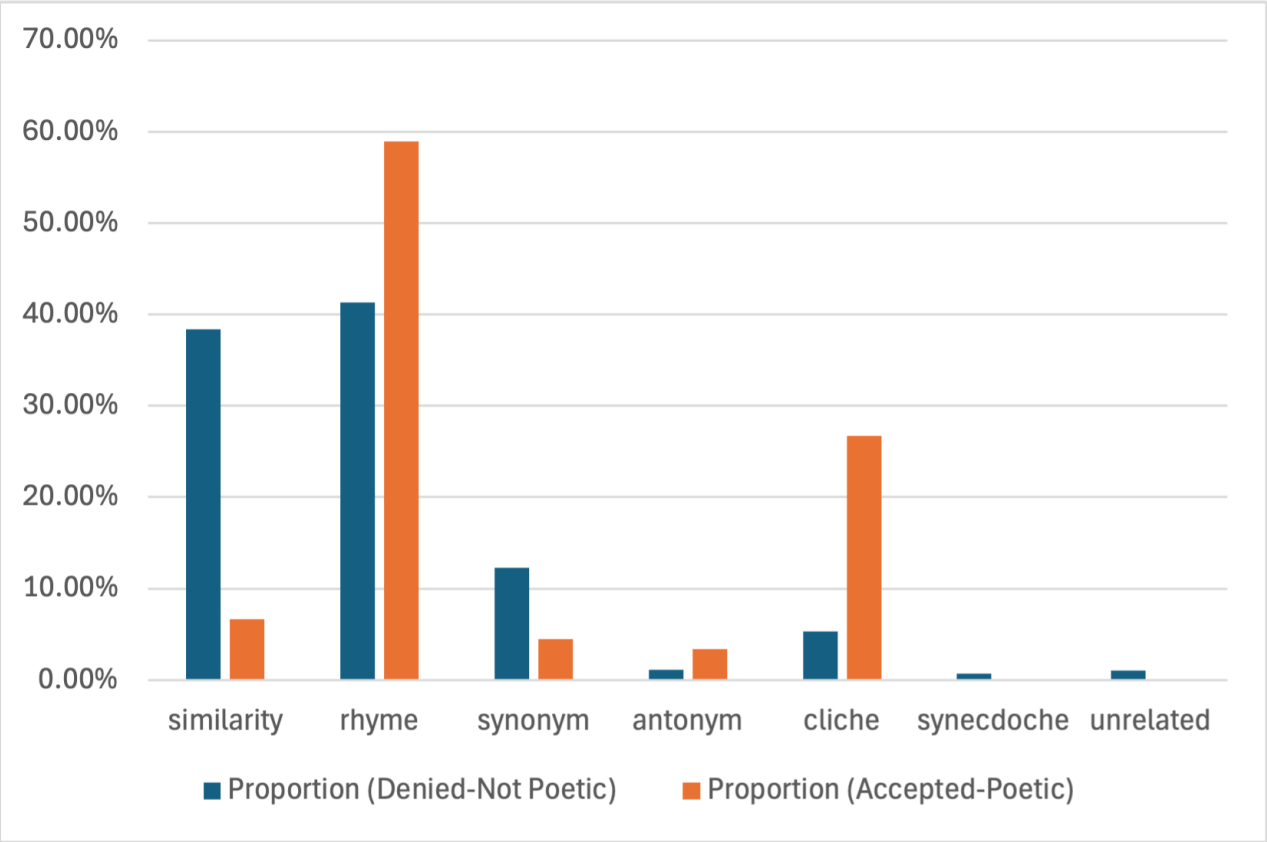

Figure 4. Analysis of the pairing results of 1000 generated couplets using Method #1 (GPT4 + ChromaDB).

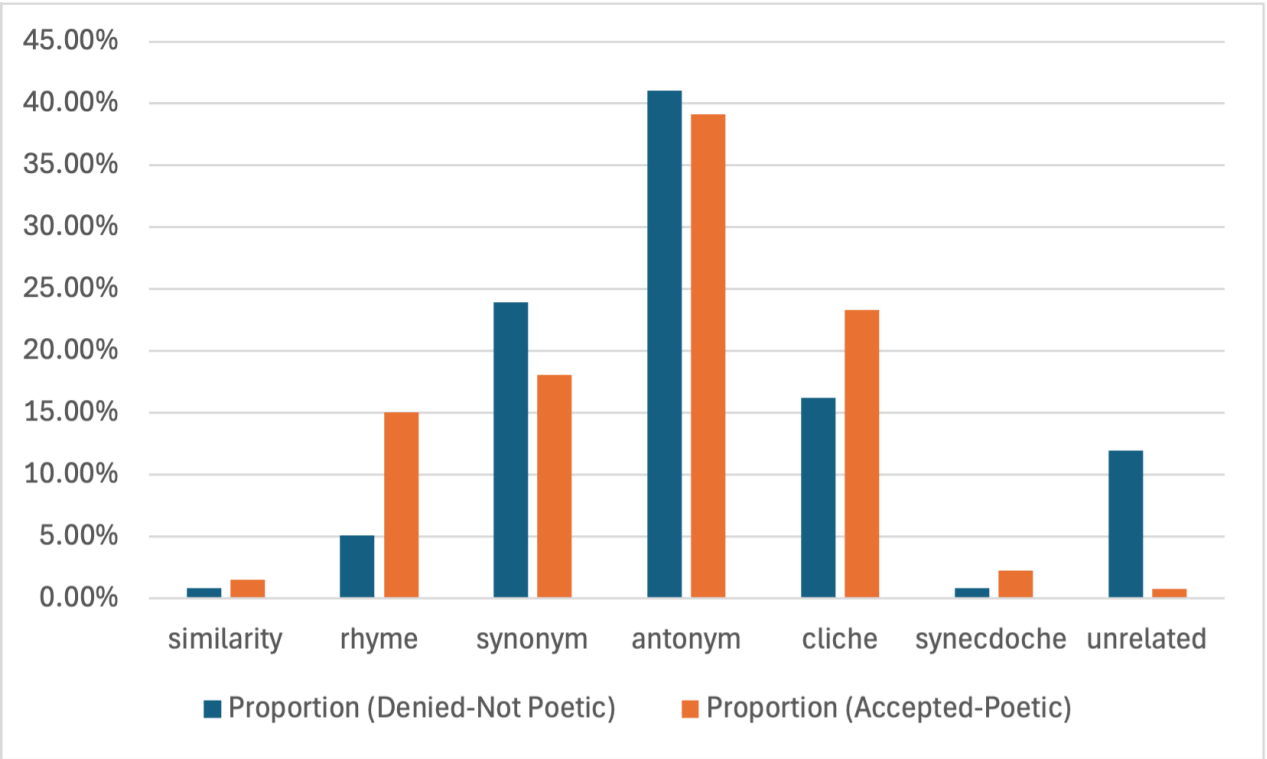

Figure 5. Analysis of the pairing results of 250 generated couplets using Method #2 (ChromaDB + GPT4).

The results (Figure 3) supported the prediction that Method #2, with AI making the pairing, would yield more poetic pairings. This is substantiated by results that point to Method #2 being almost six times more likely to generate acceptable pairings. To expedite analyzing the data, we used sampling for Method #2, looking at a quarter of the output. This may be a limitation, yet the discrepancy is so statistically different that we expect it would be observable regardless of sample size.

Figures 4 and 5 delineate the classification of the relationship between generated end words per method.

[1] Wordlist from NASPA Zyzzyva version 3.2.1 of the 2019 Collins Scrabble Words (CSW): https://www.scrabbleplayers.org/w/NASPA_Zyzzyva_Download

one more bond
one more bomb

one more crutch
one more brush

one more crude
one more crisis

one more attic
one more auto

one more basis
one more bask

one more amaze
one more amass

one more bazaar
one more razor

Figure 6. Highlights of couplets generated via Method #1.

One of the most obvious results of Method #1 was word form similarity, rhyme (including alliteration), and synonyms as dominant relationships in the output. For example, if the first end word started with the prefix *de*-, in most cases, so did the pairing. This accounts for the abundance of word form similarity (s), but we did not expect to find an even higher degree of rhyme. Similar to the cause of the prefix duplication, this method generated many alliterative pairings because so many of the nearby words started with the same letter in the embedded space of the vector database (the semantic similarity foundation also accounts for the high number of synonyms). While alliterative computational pairings can be logically accounted for, Method #1 might prove of interest and continued exploration for experimental, sound-based poets (not to mention TTS outputs in artworks). The poet lead author confesses to having a high affinity for alliteration, often using the poetic expression technique, which might have biased some of this data's acceptance rate compared to another poet's analysis. A similar surprise, but to an even greater extent, is caused by the output using rhyme, even to the word form with many pairings having similar letters, but also the same number of letters. Combined, these relationships constituted the majority of accepted couplets (~60%), but overall, the quantity of acceptable "poetic" couplets generated (albeit with much less effort) using this method was lower than manual prompting, with less than 10% compared to 15% (Experiment 1).

A qualitative sample of favorite acceptable ("poetic") output from Method #2 can be seen in Figure 6.

Turning to the results of Method #2, it is of little surprise the LLM's pairings evidenced a reliance on synonyms and antonyms to establish pairing end words. Interestingly, antonyms were the highest relationship for accepted couplets (~40%), trailed by cliché or idiomatic uses (24%), which was roughly a consistent value between the two methods, whereas Method #1 contained far fewer synonyms and antonyms as relationships (overall and in accepted couplets).

To provide more insight into the "poetic" pairings, during development of the poetry generator, we added a rationale statement—via Claude's suggestion—prompting GPT4 to explain each of its output pairings. This Chain-of-thought practice is a common way to evidence reasoning steps of AI, so we incorporated it. [22] There is certainly another small paper in further analysis of that data, but preliminary findings suggest ~18% of the time, AI is not always capable of expressing the poetic relationship it is generating.

one more is
one more was

one more nine
one more cat

one more meme
one more dna

one more ain
one more rain

one more ala
one more omega

one more mon
one more day

one more yeah
one more whisper

one more unhurried
one more constellation

Figure 7. Highlights of couplets generated via Method #2.

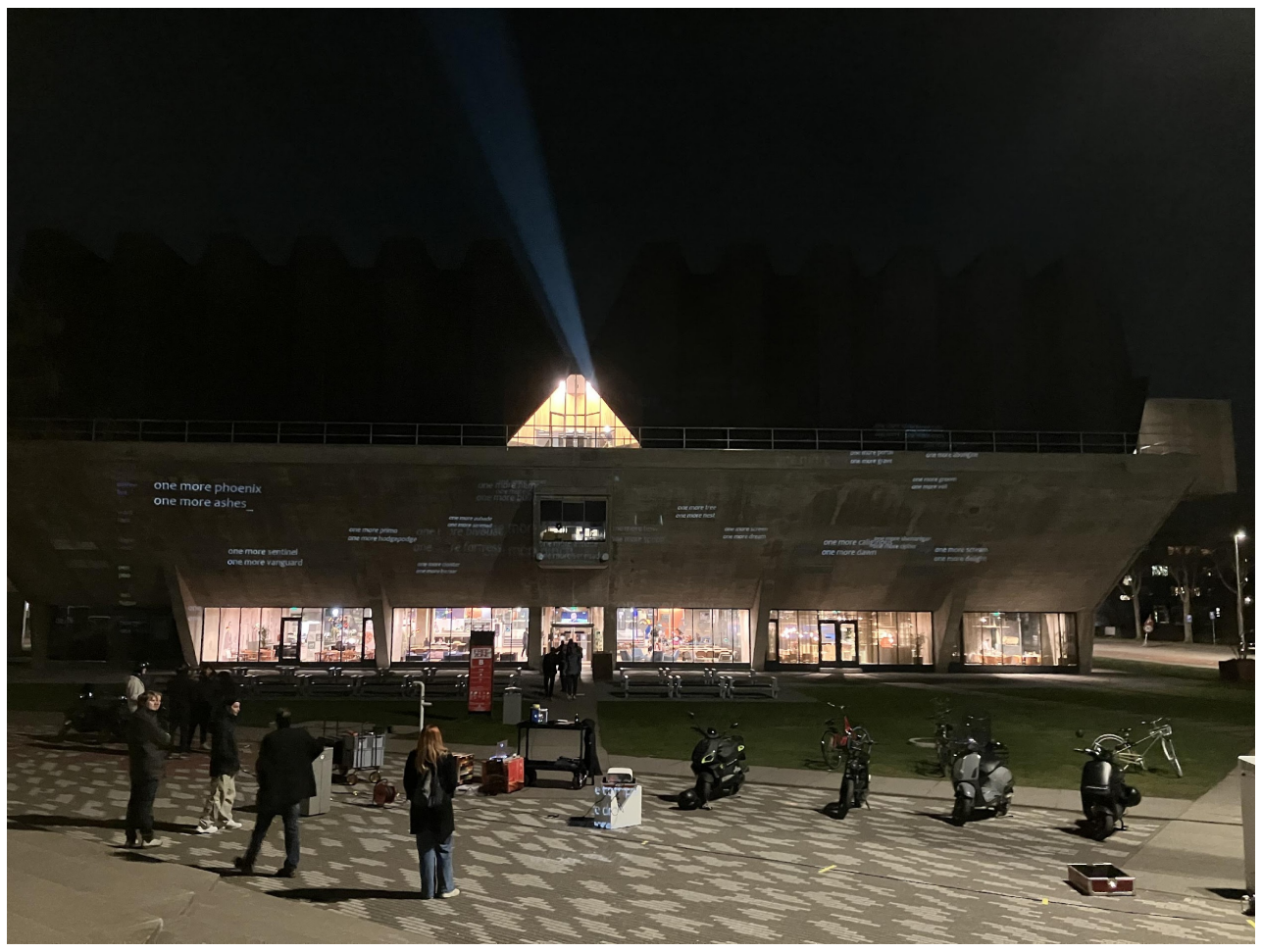

Figure 8. *resurrection 2.0* test with projection mapping, 2025.

A qualitative sample of favorite acceptable ("poetic") output from Method #2 can be seen in Figure 7.

## Artwork 2: interactive *resurrection 2.0*

In parallel to generation experimentation, the lead author developed a second iteration of *resurrection*, adding interactivity and experimenting with immersive potential using projections (Figure 8). Users can now interact with the sculpture, contribute their own pair of end words to what becomes a living poetry mural. Each user is assigned a unique color, and after they have submitted their pairing, they can instantly match the color on their mobile screen to their contributed end words, populating with a typing animation on the visualization (be it screen or projection). This is accomplished via a simple webform sent to a web server, where the text pairings are passed to the visualization application coded in Unreal. Via NDI stream technology, multiple projectors display the visualizations for variable site-specific installations. [23]

At first, the lead author felt like interactivity was needed for more engagement with the artwork, and there were multiple avenues considered (including co-creation with AI pairings), but it now seems clear that collecting solely human-generated data (hopefully generating 1000 entries) would be worth comparison to the generated couplet datasets for future study.

## Discussion and Conclusion

Our work is still in an explorative phase and should be seen as a methodological framework for artist-led evaluation of generative systems, not just an AI engineering pursuit. Below, we take the opportunity to provide a few overview reflections on limitations, our experiences, and findings with suggestions for future work.

Despite the constrained setup's advantages as an empirical methodology testing AI's generation capacity by reducing variables to the singular word level in the act of creatively pairing related end words, we are suspicious that these constraints may also be contrary to the foundation of LLMs. Given LLMs stitch together language based on a probability landscape of tokens occurring near each other, it is logical that these operations might have an issue with generation at the singular word level. Unlike humans, AI does not think in words or reliably produce words at specified quantities. This fact reinforces how LLMs only *seem* to think in language, and why humans should be suspicious that AI *knows* anything at all—certainly it does not share a human understanding of words as a basic unit of language and/or meaning. Perhaps similar experiments could be designed at the token level to see if there might be an even more conducive way for the LLMs to generate poetic, linguistic relations. We also want to forefront using single rater evaluation as a limitation and foresee experiments using multiple third-party raters (including AI) for future work.

There are additional considerations to Experiment 2 that could also be limitations in terms that the values used for relatedness in ChromaDB selections could be altered and optimized. Other n_results distance values are worth exploring, as well as other generations of the dataset, even variations of prompts. We plan to expand this research by fully analyzing the output and producing two datasets of each method at 1000 generations, so they can be compared between the same generation method as well as alternative methods. We are still exploring other methods of generation and pairing that might lead to even more fruitful findings.

Furthermore, infinity is obviously hyperbole when you consider the state of AI and the limits of a language dataset. There are approximately half a million words in English (450,000 words in *Webster's Third Dictionary* and 500,000 in *Oxford English Dictionary*). Therefore, to be precise, it's a quadratic rather than an exponential capacity, as there are certainly far fewer than 250 billion combinations, given that not all pairings would be suitable. But a figure of this order is still massive. Even at half (to compensate for unsuitable pairings), 125 billion combinations, if generated daily, would take 342.5 million years to reach a limit. If you wanted to finish in 80 years (a figure representative of the human lifespan), it would take over 4 million a day, which is approximately 50 generated entries per second. These calculations, while primitive and done with the assistance of AI, make the lead author realize the quantitative goal and speed of one couplet per day (*resurrection 1.0*) could be dramatically increased for an artwork that tests the very limits of word combinations. Furthermore, the exponentiality can be reintroduced with an experiment of a longer linguistic structure, especially one creating new compound words or line/phrases, but this becomes proportionately harder to analyze and assess.

While the quantitative possibilities are impressive, the poetic semantic relationship between the generated end words is less. The machine (even with artificial intelligence) is far less capable of succeeding in this minimalistic poetic act than humans, even ones with limited experience in poetry. With more data collection of the human-generated datasets from the presentation of the interactive *resurrection*, we plan on validating these findings with comparative evidence in continuation of the research. But for the moment, we find creativity and poetry to still reside mostly "naturally" in human minds, especially for pairing words with a relationship that can be considered poetic as opposed to pairings with specified relationships like rhyme, synonyms, or antonyms. So, literary writers are far from being replaced; in fact, with everyone so loosely wielding LLMs, writers are more irreplaceable than ever to put pressure and proliferate the meaning of words (which are living and changing). Meaning has always been intrinsically connected to impermanence or mortality. Our fears of being replaced by machines are simply an extension of our mortal turmoil, with AI being the contemporary face of that ancient threat. Language is one of the only tools we have invented that seems to alleviate the circumstance of mortality, enabling our thoughts and writing to vastly exceed our lifespan. Authors are resurrected when we read them, and poetry is resurrected from output whenever a poet puts their hands on the data.

AI is certainly a tool to keep testing for poetry, one that can aid in hunting and mixing language, opening up many new possibilities, even if its pairings are less inspiring (sometimes amazing outliers are generated, but as a poet, the lead author finds the vast majority inferior to his own pairings). Poetic pairings come from experience, which again brings up colloquial usage–something an LLM should be statistically privy to via its online training dataset, but is incapable to fully grasp. An example of this is "apples" and "oranges"; while the machine can retrieve these are fruits, and that there is a popular expression relating the two, it certainly doesn't understand that this cliché expression can be mocked or that a substitution of "orangutans" for oranges recalls the idiom but takes it into new poetic territory. While this is just an arbitrary example, these observations recall Bense's criticisms of generators of artificial poetry, which only create "*unwillkürlicher Bedeutungswelt* [worlds of meaning, involuntarily (without intent)]." [24] We are not so sure this will hold as AI continues to advance at remarkable rates. We would like it to generate better experimental poetry. Given humanity's history of playing with language, we are confident humans will keep playing with LLMs, and poets—just like they do with language—will imbue new ways to create into Gen AI as they, in turn, use it to create in new ways.

Nitpicking poetry was never the point; our goals relate to incorporating this technology meaningfully into literary practices and experiences. We are attracted to research new computationally creative technology, such as AI, to re-engage with poetry, creating new horizons for reading and/or writing, "multi-modal interfaces, [which] challenge traditional ways of understanding how, where, and with what capacities we read [/write].” [25] The lead author plans to use the work of *resurrection* as a first case study to generate a framework for producing more works combining poetry and computational creativity. Instead of witnessing these developments in a state of thanatophobia, he shares Richard Lanham’s belief that “electronic expression has come not to destroy the Western arts and letters, but to fulfil [/resurrect] them.” [26]

## Acknowledgements

Experiment #2 couplet generator relied on coding support from Reimer Rigtering. Additional technical support for the art installation versions were collaborations with Koen Wösten as well as Arend-Jan Krooneman, Luuk Goossen, and Roland van Roijen from the TU Delft NewMedia Centre. The artwork of Sophia Bulgakova and Katarina Petrovic has also certainly influenced this work. Dr.ir. Kim Lompe assisted with German translations.

## Author Biographies

Vincent A. Cellucci is a poet and PhD candidate at Leiden University. His research involves using AI to continue to push the envelopes of experimental writing, publishing, performance, interactivity, and exhibitions in the worlds of education and fine arts.

Peter van der Putten is an assistant professor in AI and creative research at LIACS, Leiden University, heads up the AI Lab at Pegasystems, and is an independent AI artist and curator.